\documentclass[10pt, a4paper]{article}

\usepackage[]{lrec-coling2024} 
\usepackage[]{authblk}
\usepackage{graphicx}
\usepackage{tabularx}
\usepackage{soul}

\usepackage{microtype}
\usepackage{graphicx}
\usepackage{multirow}
\usepackage{tikz}
\usepackage{tikz-qtree}
\usepackage{amsmath}
\usepackage{dashbox}

\usepackage{fbox}
\usepackage{longfbox}
\usepackage{booktabs}
\usepackage{titlesec}
\usepackage{pgfplots}
\usepackage{tikz}
\usepackage{gb4e}
\usepackage{xcolor}
\usepackage{blindtext}
\noautomath

\newcommand*{\email}[1]{%
    \small\href{mailto:#1}{#1}\par
    }

\title{SPLICE: A \underline{S}ingleton-Enhanced \underline{P}ipe\underline{LI}ne for \underline{C}oreference R\underline{E}solution}

\name{Yilun Zhu$^\spadesuit$, Siyao Peng$^\heartsuit$, Sameer Pradhan$^{\clubsuit\diamondsuit}$, Amir Zeldes$^\spadesuit$\\}

\address{$^\spadesuit$ Department of Linguistics, Georgetown University \\ 
         $^\heartsuit$ MaiNLP, Center for Information and Language Processing, LMU Munich \\ 
         $^\clubsuit$ Linguistic Data Consortium, University of Pennsylvania \\ 
         $^\diamondsuit$ cemantix.org \\
        \email{\{yz565, amir.zeldes\}@georgetown.edu}, \email{siyao.peng@lmu.de}, \email{pradhan@cemantix.org}\\}

\abstract{
Singleton mentions, i.e.~entities mentioned only once in a text, are important to how humans understand discourse from a theoretical perspective. However previous attempts to incorporate their detection in end-to-end neural coreference resolution for English have been hampered by the lack of singleton mention spans in the OntoNotes benchmark. This paper addresses this limitation by combining predicted mentions from existing nested NER systems and features derived from OntoNotes syntax trees. With this approach, we create a near approximation of the OntoNotes dataset with all singleton mentions, achieving $\sim$94\% recall on a sample of gold singletons. We then propose a two-step neural mention and coreference resolution system, named SPLICE, and compare its performance to the end-to-end approach in two scenarios: the OntoNotes test set and the out-of-domain (OOD) OntoGUM corpus. Results indicate that reconstructed singleton training yields results comparable to end-to-end systems for OntoNotes, while improving OOD stability (+1.1 avg. F1). We conduct error analysis for mention detection and delve into its impact on coreference clustering, revealing that precision improvements deliver more substantial benefits than increases in recall for resolving coreference chains.
\\ \newline  }

\begin{document}

\maketitleabstract

\section{Introduction}
Coreference is a linguistic phenomenon in which two or more expressions (also known as mentions) in a text refer to the same entity (e.g.~\textit{the Vice President ... She}).
To correctly cluster mentions, the first step is identifying referring expressions, candidates for repeated reference in context.
Such mentions include both coreference markables, which are expressions part of a coreference chain, and singletons, which could be referred back to but are not involved in any coreference relations in the given text.
From a theoretical linguistic perspective, all mentions are important for coreference resolution because humans understand discourse and entity coherence based on the competing available options \cite{grosz-etal-1995-centering}. From an empirical perspective, both markables and singletons are important components in the data distribution for cluster linking, with coreference markables corresponding to true positives and singletons corresponding to true negatives \cite{kubler-zhekova-2011-singletons}, while all mentions can be used to improve coreference markable boundary detection.

In recent years, end-to-end \cite{lee-etal-2017-end, lee-etal-2018-higher} and sequence-to-sequence \cite{bohnet-etal-2023-coreference} approaches have demonstrated superior performance compared to rule-based \cite{raghunathan-etal-2010-multi, lee-etal-2013-deterministic} and entity-based neural approaches \cite{wiseman-etal-2015-learning,clark-manning-2015-entity,clark-manning-2016-deep,clark-manning-2016-improving}. 
Despite progress achieved by deep learning models, the proposed solutions diverge from discourse theory by not considering all mention candidates, particularly singletons, when learning coreference linking, which also results in limited interpretability of existing models. The major reason that this issue has been overlooked is that the dataset used for training most coreference resolution systems, OntoNotes \cite{pradhan-etal-2013-towards}, lacks singleton annotations.\footnote{There were two main reasons for not annotating singletons in OntoNotes: i) Annotating singletons would have increased the annotation effort significantly; therefore, a trade-off had to be made; ii) Inter-annotator agreement on whether or not a text span is referential was also relatively low.}

\begin{figure*}[t!hb]
    \centering\small
    \begin{tikzpicture}[scale=0.75,every node/.style={scale=0.9}]
    \tikzset{edge from parent/.style=
    {draw,
    edge from parent path={(\tikzparentnode.south)
    -- +(0,-5pt)
    -| (\tikzchildnode)}}}
    \Tree
    [.TOP 
        [.S-IMP [.SBAR-ADV [.IN If ]
                          [.S [.\dbox{NP-SBJ-1} [.PRP you ] ]
                             [.VP [.VBD were ]
                                 [.VP [.VBN saved ]
                                     [.\dbox{NP} [.-NONE- *-1 ] ]
                                     [.PP-CLR [.IN from ]
                                             [.\node[draw]{NP}; [.DT the ]
                                                 [.NN lion ] ] ] ] ] ] ]
                [.VP [.VB do ]
                    [.RB n't ]
                    [.VP [.VB start ]
                        [.S [.\dbox{NP-SBJ} [.-NONE- *-2 ] ]
                           [.VP [.VBG wishing ]
                               [.S [.\dbox{NP-SBJ} [.-NONE- *PRO* ] ]
                                  [.VP [.TO to ]
                                      [.VP [.VB hunt ]
                                          [.\node[draw]{NP}; [.PRP it ] ] ] ] ] ] ] ] ]
                [.FRAG [.\node[draw]{NP}; [.NNP Mona ]
                          [.NNP Lisa ] ] ] 
        ] 
    ]

    \end{tikzpicture}
    \caption{An example of the utilization of a syntax tree for the extraction of mentions. The solid box signifies that the NP is a candidate for coreference linking in OntoNotes while the dashed box indicates that the NP is not categorized as a mention.}
    \label{fig:tree}
\end{figure*}
Previous work on recovering singletons in the data initially used gold syntax annotations from OntoNotes to develop a rule-based algorithm. \citet{raghunathan-etal-2010-multi} extracted pronouns and maximal NP projections and incorporated a post-processing step that employed a set of rules to filter out mentions that didn't align with the annotations, such as numeric mentions. Such approaches have been used as a preprocessing step for several coreference systems \cite{wiseman-etal-2015-learning, wiseman-etal-2016-learning}. 
Another strategy involves parsing syntax trees to identify all NPs from the corpus \cite{clark-manning-2015-entity, clark-manning-2016-improving}, which was frequently used before the advent of end-to-end systems. 
The third method aims to generate silver singletons for the corpus. \citet{recasens-etal-2013-life} proposed a lifespan model to make distinctions between singleton mentions and coreference markables. More recently, \citet{toshniwal-etal-2021-generalization} proposed a data augmentation strategy to extract silver mentions (`pseudo-singletons') from a mention detector trained on OntoNotes coreference markables. 
\citet{zhu-etal-2023-incorporating} demonstrated that gold singletons and mention-based features can improve coreference scores on OntoGUM, an out-of-domain (OOD) dataset following the OntoNotes scheme.

However, these methods exhibit certain limitations. While extracting NP subtrees can achieve a high recall in singleton detection, it concurrently generates a large number of precision errors (spans that are not valid mentions). For example, the span \textit{you} in Figure \ref{fig:tree} is a valid NP but is not a mention candidate for pair matching as it is a {\em generic} you. 
By contrast, the second method is trained to pick up OntoNotes mentions, but falls short due to two reasons: (1) the system is biased towards mentions that resemble coreference markables in OntoNotes, missing atypical ones with semantic and syntactic disparities; (2) evaluating performance of the mention detector is challenging without any gold singletons, meaning we do not know how its performance is impacting downstream coreference scores. Finally, we note that in realistic settings, applications may want access to all entities mentioned in a text, including singletons, meaning their comprehensive detection is desirable. 

Based on the necessity of singletons for context understanding and the existing problems of previous research, our paper aims to extract near-gold singletons using datasets with singleton annotations. We demonstrate how to effectively employ these in coreference systems to improve in/out-of-domain performance. The contributions\footnote{
The code for training the mention classifier \& the coreference pipeline and data of OntoNotes singletons are publicly available at \url{https://github.com/yilunzhu/splice}.} include:
\begin{itemize}
\item A mention detection classifier that extracts mentions from syntactic structures and achieves a recall of $\sim$94\% on the OntoNotes development set. 
\item A near-gold singleton annotated version of OntoNotes.
\item A pipeline-based neural coreference system, named SPLICE, using singletons, yielding results on par with the end-to-end approach in-domain and a +1.1 boost OOD.
\item Evaluation at different precision and recall levels for mention detection and analysis of the effect of singletons on coreference linking.
\end{itemize}

\section{Related Work}
\paragraph{Mention Detection}
As a preprocessing step, mention detection is an important component of coreference resolution. Most neural approaches implement mention detection as part of an end-to-end system. The widely-used pre-neural OntoNotes mention detector employed a rule-based system to extract pronouns and maximal NP projections given gold syntax trees \cite{raghunathan-etal-2010-multi}, while the end-to-end approach from \citet{lee-etal-2017-end, lee-etal-2018-higher} detected markables directly during pair matching. \citet{yu-etal-2020-neural} compared three neural approaches for scoring markable spans and proposed using a biaffine model, which achieved a high recall on mention detection.

\paragraph{Coreference Resolution}
Recent neural coreference resolution systems have achieved great improvements. The end-to-end approach \citep{lee-etal-2017-end} jointly learns mention detection and coreference pair scoring and achieved SOTA scores on the OntoNotes test set before several extensions were proposed. \citet{lee-etal-2018-higher, kantor-globerson-2019-coreference} improved span representations to improve pair matching. \citet{joshi-etal-2020-spanbert} added better pre-trained language models to gain additional score boosts. \citet{wu-etal-2020-corefqa} adapted a question-answering framework into the task and improved both coreference markable span detection and coreference matching scores. \citet{dobrovolskii-2021-word} also improved performance by initially matching coreference links via words instead of spans. Recently, \citet{bohnet-etal-2023-coreference} proposed a sequence-to-sequence paradigm to predict mentions and links jointly. However, none of these models consider singletons for coreference linking.

\section{Nominal Phrase Extraction}
Mentions are typically manifested as noun phrases (NPs) in syntactic structures.\footnote{Some coreference guidelines also consider verbs referred back to by NPs (which are then a fraction of all mentions) as mention candidates.} However, due to the intricate nature of NPs, whose recognition entails prepositional phrase (PP) attachment disambiguation and diverse sentence structure analyses, a considerable portion of NPs is excluded from consideration as valid mentions for subsequent coreference resolution according to the annotation guidelines of OntoNotes,\footnote{\url{https://catalog.ldc.upenn.edu/LDC2013T19}} such as nested mentions inside proper names, generic \textit{you}, expletives, adjectives and non-proper nouns within pre-modifiers, etc. For example, the sentence in Figure \ref{fig:tree} contains 7 NPs. Three of them are syntactic traces, marked by \textit{-NONE-} in the node, and one is a pronominal phrase without phonological content, marked as \textit{*PRO*}, both of which are discarded during the pre-processing step. There are three possible coreferential mentions: \textit{the lion}, \textit{it}, and \textit{Mona Lisa}. By contrast, the NP \textit{you}, cannot be considered a potential mention during linking since generic \textit{you} is not annotated as a coreference markable in OntoNotes. To prevent the inclusion of ``invalid'' mentions during the coreference linking stage, excluding these NPs from the list of potential mention candidates is helpful.

\subsection{Dataset Preparation for NPs} \label{sec:data_prep}
In the coreference layer, OntoNotes does not have singleton annotation. Although the corpus also has named entity annotations, these cannot solve the problem because only flat named entities are annotated (no non-named or nested mentions). However, not only flat named entities, but also nested, named, and non-named entities (=coreference markables + singletons) are candidates for coreference linking according to OntoNotes annotation schema. When constructing the corpus, annotators were offered gold pronouns and NPs as coreference markable candidates, meaning gold syntax trees could be utilized to recall near-gold singletons. 
The precision of mapping syntactic NPs to singletons could be studied using data with both annotation types.
To the best of our knowledge, only ARRAU \cite{poesio-etal-2018-anaphora} and OntoGUM \cite{zhu-etal-2021-ontogum} meet these requirements.\footnote{ARRAU annotates more mentions compared to OntoNotes, including non-referring expressions, such as \textit{on [the other hand]}.} Among the four genres in the ARRAU corpus, the RST news genre consists of the subset of the Penn Treebank (PTB, \citealt{marcus-etal-1993-building}) that was annotated for discourse relations in the RST treebank \cite{carlson-etal-2001-building}. Since OntoNotes overlaps the same subset of PTB, ARRAU can be used to compare singleton annotations with PTB constituent trees, though its coreference scheme differs from OntoNotes. OntoGUM, designed to follow the OntoNotes annotation scheme and featuring 12 genres, is a second option, though it has predicted constituent trees converted from gold dependency structures. Thus, OntoGUM can provide OOD data to make the classifier more robust across genres/datasets. OntoGUM V9 is utilized in this paper. We use the two datasets to create a classification model to map gold NPs to near-gold singletons.
Since RST documents overlap OntoNotes, we rearrange document splits to facilitate downstream mention classification and coreference: train/test: 265/148.\footnote{We release our re-split together with our model.}

\begin{table}[t!]
    \centering\small
    \begin{tabular}{l|c|c|c|r}
        \toprule
        Category & P & R & F1 & Num. \\
        \hline
        \rule{0pt}{2ex}0 & 0.92 & 0.84 & 0.87 & 8,087 \\
        1 & 0.95 & \textbf{0.97} & 0.96 & 23,877 \\
        Micro Avg & 0.94 & 0.94 & 0.94 & 31,964 \\
        \bottomrule
    \end{tabular}
    \caption{Results of the XGBoost NP classifier on the test set (new data split excluding OntoNotes test documents) of ARRAU. \textbf{1} denotes the NP is a mention and \textbf{0} presents the opposite.}
    \label{tab:classifier}
\end{table}

\subsection{Mention Classification}
Let $I=\{1,...,i\}$ be the number of NPs and pronouns within a document. This classification task aims to distinguish mentions that can potentially have coreference relationships (referring expressions) from other NPs.

We use ARRAU's RST portion with the above split for training and evaluation. First, all pronouns and NPs are extracted from the gold syntax trees so that each span can be allocated a label based on ARRAU's coreference layer annotations. Pronouns identified through a rule-based function and NPs annotated within ARRAU are assigned positive labels. Other NPs are assigned negative labels.

For effective training of an NP classifier, relying solely on information within mentions spans is insufficient: NP that would be mentions in one context may not be when nested in another phrase. Furthermore, the variable count of parent and child spans in each phrase introduces challenges if we want to avoid handpicked features. To address these problems, we introduce a set of generic features that describe each NP span and its graph position. The features for each NP consist of two primary components. The first set of features encompasses mention-based features of the current NP, its parent phrases, and child phrases. These features include parts-of-speech tags, the usage of prepositions, definite markers, grammatical roles, adverbial tags, etc. Additionally, we extract features from other NPs that overlap with the current one, considering features like their relative positions or hierarchical levels among other NPs, as well as the largest and smallest interactive NP spans. We select the XGBoost classifier \cite{Chen:2016:XST:2939672.2939785} for the NP classification task and evaluate its performance on the ARRAU test set. We train the \texttt{gbtree} booster with a \texttt{learning\_rate} set at 0.1. Due to some disparities between OntoNotes markables and PTB NPs, a small portion of mentions cannot be extracted via syntax structures, for example, compound modifiers such as `[Hong Kong] government.' In such cases, we introduce a post-processing step to recover non-NP singletons.

Table \ref{tab:classifier} shows that the classifier performs well on ARRAU, demonstrating the model's capability to map PTB tree patterns to distinguish potential coreference markables. In particular, the model excels in recall of coreference markables. A recall-focused model offers the advantage of generating a substantial pool of candidates for potential linking, affording the coreference resolver a secondary opportunity to eliminate non-mention spans.

\begin{table}[t!]
    \centering\small
    \begin{tabular}{l|c|c|c}
        \toprule
        Dataset & P & R & F1 \\
        \hline
        \rule{0pt}{2ex}ARRAU & 28.15 & 97.78 & 44.35 \\
        OntoNotes & 39.46 & 91.65 & 55.16 \\
        \bottomrule

    \end{tabular}
    \caption{Results of coreference markables on ARRAU and OntoNotes test captured by the NP classifier.}
    \label{tab:classifier_eval}
\end{table}

Additionally, we want to evaluate the classifier's ability to identify coreference markables. Table \ref{tab:classifier_eval} presents the scores for coreference markables as captured by the classifier both in-domain (ARRAU) and OOD (OntoNotes). Given the minor disparities in annotation and genre between the two datasets, the classifier performs well (here we focus on the recall score) in distinguishing mentions from NPs. The predictions rendered by the classifier are utilized in training the mention detector and is available at the public GitHub link introduced in footnote 2.

\begin{figure*}[t!]
\centering
\includegraphics[width=\textwidth]{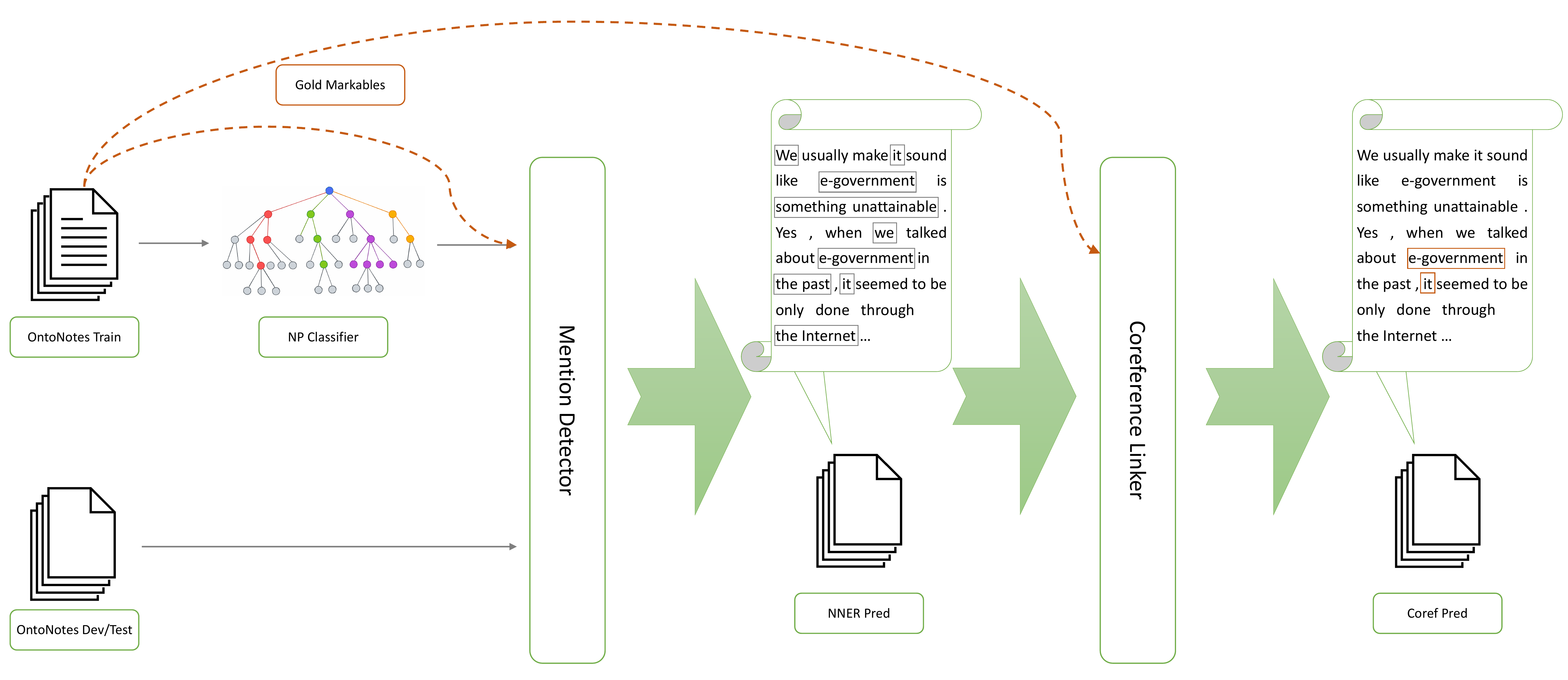}
\caption{The Pipeline of the Two-step Coreference System Using Singletons. Gold markable spans are leveraged for training mention detection and coreference linking to enhance alignment with the OntoNotes annotation schema.}
\label{fig:pipeline}
\end{figure*}

\section{Coreference Pipeline}
With the enhanced OntoNotes data in hand, we build a training pipeline for coreference inference.

\subsection{Dataset}
OntoNotes V5.0 
\citep{pradhan-etal-2013-towards} contains 1.6M tokens annotated for coreference, with a test set comprising 348 documents with $\sim$170K tokens. We also use OntoGUM V8.0 \cite{zhu-etal-2021-ontogum} as an OOD dataset to evaluate the model's generalization performance. OntoGUM's test set includes 24 documents with $\sim$22K tokens, following the same coreference annotation scheme as OntoNotes.

\subsection{Training} \label{sec:training}

\paragraph{Mention detector} We use the classifier trained on ARRAU to predict positive and negative labels within the OntoNotes training dataset. Then, we take the union of the classifier's outputs and gold coreference markables from the OntoNotes train set. This data serves as input for the mention detector. We realize that the mention span detection task closely resembles the nested named-entity recognition (NNER) task, with the key distinction being that mention detection does not require entity types. Thus, we train a SOTA NNER system \cite{tan2021sequencetoset} on the union data, which we will use to predict mentions at test time, as shown in Figure \ref{fig:pipeline}. 

\begin{table*}[t!b]
    \centering\small
    \begin{tabularx}{\textwidth}{X|>{\hsize=1.2\hsize}X|>{\hsize=1.2\hsize}X|>{\hsize=.5\hsize}X}
    \toprule
    Data & Precision & Recall & F1 \\
    \midrule
    OntoNotes-dev & 37.84 \hspace{2em} (18,321/48,419) & 95.64 \hspace{2em} (18,321/19,156) & 54.22 \\
    OntoGUM-test & 37.75 \hspace{2em} (19,018/50,736) & 96.23 \hspace{2em} (19,018/19,764) & 54.23 \\
    OntoGUM-test & 37.21 \hspace{2em} (\enspace 2,439/\enspace 6,554) & 91.66 \hspace{2em} (\enspace 2,439/\enspace 2,661) & 52.94 \\
    \bottomrule
    \end{tabularx}
    \caption{Mention detection performance on OntoNotes dev/test set and OntoGUM test set.}
    \label{tab:md_perf}
\end{table*}

\noindent
The performance results are presented in Table \ref{tab:md_perf}. The mention detector demonstrates a high recall rate, identifying approximately 96\% of coreference markables in both the validation and test sets of OntoNotes. It reveals that it effectively captures a significant portion of the relevant information according to OntoNotes' guidelines. However, the precision score is comparatively lower at $\sim$37.8\%.
Since there are approximately twice as many mentions as coreference markables \cite{zhu-etal-2021-ontogum}, this suggests an estimated count of around 40K gold mentions in OntoNotes dev and test sets each. 
However, the currently extracted mentions contain nearly 20\% ($10\text{K}=48\text{K}-19\text{K}*2$) of incorrectly predicted spans. Though the mention classifier and the mention detector achieve the best results in extracting near-gold singleton spans from OntoNotes, the low precision score for coreference markables indicates that a significant number of identified mentions may distract the coreference linker.

The extracted mentions are then used in the coreference model training process, which learns to identify and link valid mentions implicitly. In OOD evaluation, it is also observed that the mention detector produces high-quality mentions with a recall of nearly 92\%.

\paragraph{Coreference model}
We use the end-to-end (e2e) model with SpanBERT-large embeddings \cite{joshi-etal-2020-spanbert} as our baseline model. The baseline considers all span possibilities during coreference linking. As in (\ref{formula:baseline}), it uses a feed-forward network to compute a markable score for each possible token span, represented by a concatenation of four vectors: token embedding of the start token and the end token of the span, attention-based head embeddings, and meta information (such as genres, gold speaker information) shown in (\ref{formula:span}). The neural network calculates a mention score for each span. It then keeps a fixed number of spans with top scores for coreference clustering. However, this pruning method may exclude correct markable spans and singletons with lower scores during inference.
For this reason, our proposed pipeline diverges by exclusively training on the union of gold coreference markables and positive outputs from the mention detector. We then assign identical mention scores to all spans, ensuring that the likelihood of span mentions does not influence the model's coreference linking decisions. As illustrated in (\ref{formula:new}), we utilize a trainable parameter $w_m$ for the markable score and employ hyperparameter tuning to find the best alignment with coreference clustering.

\begin{exe}
    \ex \[\text{Baseline: } s_m = \texttt{FFNN}_{\texttt{m}}(g_i)\] \label{formula:baseline}
    \ex  \[g_i = [x_{\texttt{start(i)}}, x_{\texttt{end(i)}}, \hat x_{\texttt{i}}, \phi (i)]\] \label{formula:span}
    \ex \[\text{Ours: } s_m = w_m\] \label{formula:new}
\end{exe}

\noindent
To maintain consistency in training experiments, we keep other hyperparameters at the same values as the baseline model, mitigating the impact of external factors. The proposed pipeline for training and inference is outlined in Figure \ref{fig:pipeline}.

\subsection{Inference}
The evaluation of the coreference resolution task requires plain text as input, meaning mention spans and gold syntax trees cannot be used at test time. Consequently, as shown in Figure \ref{fig:pipeline}, inference is divided into two steps: First, the mention detector (based on an NNER system) reads the plain input and generates nested mentions. Second, these predicted mention spans are provided as input to the coreference model, which constructs coreference chains based on them.

\begin{table*}[t!b]
    \centering\scriptsize
    \begin{tabular}{l|ccc|ccccccccc|c}
    \toprule
    \multicolumn{1}{l|}{\multirow{2}{*}{}} & \multicolumn{3}{c|}{Mention Detection} & \multicolumn{3}{c}{MUC} & \multicolumn{3}{c}{B$^3$} & \multicolumn{3}{c|}{CEAF$_{\phi4}$} & \multicolumn{1}{c}{\multirow{2}{*}{Avg. F1}} \\
    \multicolumn{1}{l|}{} & P & R & \multicolumn{1}{c|}{F1} & P & R & F1 & P & R & F1 & P & R & F1 &  \\
    \midrule
    \citet{joshi-etal-2020-spanbert} &  89.1 & 86.5 & 87.8 & 85.8 & 84.8 & 85.3 & 78.3 & 77.9 & 78.1 & 76.4 & 74.2 & 75.3 & 79.6 \\
    \rule{0pt}{2ex}\rule{2ex}{0ex}\texttt{Ours+MD} & 88.8 & 87.3 & 88.1 & 85.6 & 84.5 & 85.1 & 78.8 & 77.0 & 77.9 & 75.8 & 74.4 & 75.1 & 79.4 \\
    \rule{0pt}{2ex}\rule{2ex}{0ex}\texttt{Ours+MD+GM (upperbound)} & 90.9 & 91.3 & 91.1 & 87.9 & 88.6 & 88.3 & 81.4 & 82.7 & 82.0 & 80.3 & 79.9 & 80.1 & 83.5 \\
        
    \bottomrule
    \end{tabular}
    \caption{Results on OntoNotes test set. \textit{MD} denotes the model uses predictions from the mention detector; \textit{GM} indicates the model uses gold coreference markables.}
    \label{tab:res_on}
\end{table*}

\begin{table*}[t!b]
    \centering\scriptsize
    \begin{tabular}{l|ccc|ccccccccc|c}
    \toprule
    \multicolumn{1}{l|}{\multirow{2}{*}{}} & \multicolumn{3}{c|}{Mention Detection} & \multicolumn{3}{c}{MUC} & \multicolumn{3}{c}{B$^3$} & \multicolumn{3}{c}{CEAF$_{\phi4}$} & \multicolumn{1}{|c}{\multirow{2}{*}{Avg. F1}} \\
    \multicolumn{1}{l|}{} & P & R & \multicolumn{1}{c|}{F1} & P & R & F1 & P & R & F1 & P & R & F1 &  \\
    \midrule
    \citet{joshi-etal-2020-spanbert} &  86.0 & 70.6 & 77.5 & 80.0 & 68.1 & 73.6 & 67.9 & 60.5 & 64.0 & 68.6 & 50.5 & 58.2 & 65.3 \\
    \rule{0pt}{2ex}\rule{2ex}{0ex}\texttt{Ours+MD} & 85.3 & 73.5 & 78.9 & 78.8 & 70.6 & 74.5 & 66.5 & 63.5 & 64.9 & 68.3 & 52.0 & 59.0 & 66.4 \\
    \rule{0pt}{2ex}\rule{2ex}{0ex}\texttt{Ours+GS (upperbound)} & 90.8 & 74.8 & 82.0 & 84.8 & 72.4 & 78.1 & 74.2 & 65.6 & 69.6 & 75.7 & 55.6 & 64.2 & 70.8 \\
    \bottomrule
    \end{tabular}
    \caption{Results on OntoGUM test set. \textit{GS} indicates that our model uses gold singletons.}
    \label{tab:res_og}
\end{table*}

\section{Experiments}
\subsection{Experiments Setup}

Following \citet{tan2021sequencetoset}, the mention detector uses BERT-base \cite{devlin-etal-2019-bert} as the base model to train the system.
All coreference clustering experiments use Pytorch and the pre-trained SpanBERT large \cite{joshi-etal-2020-spanbert} model from HuggingFace\footnote{\url{https://huggingface.co/}} for token representations. 
Both mention detection and coreference linking experiments are conducted on Nvidia RTX A6000 GPUs with 64GB RAM.

\subsection{Results}
\paragraph{In-domain}

Table \ref{tab:res_on} presents the results of our proposed pipeline and the baseline model on the OntoNotes test set. When using the predictions from the mention detector (\texttt{Ours+MD}), our model yields comparable mention spans and achieves a comparable average F1 score to the baseline model (79.4 vs. 79.6). This indicates our model can effectively learn to resolve coreference even when provided with imperfect input. Additionally, the results suggest that with recent improvements in nested NER systems, a sufficient number of coreference markables can be captured, making the end-to-end architecture not significantly superior to our pipeline-based system.
More than that, our pipeline-based model can output all entities: singletons and coreferring chains.

Using both gold coreference markables and predicted mentions (\texttt{Ours+MD+GM}) as inputs represents an upper bound for our pipeline-based system. Our model achieves an average F1 score of 83.5 in this scenario, marking a nearly 4-point increase over the baseline. This substantial gap indicates that, although the mention detection module generates some incorrect spans (precision errors), the coreference clustering module can generally construct correct clusters from the provided spans.

These results demonstrate that our system is a robust approach to coreference resolution in-domain. We further assess how stable OOD is below.

\paragraph{OOD performance}
Table \ref{tab:res_og} compares the baseline model and our system for OOD evaluation on the OntoGUM test set, which includes 12 written and spoken genres, especially challenging ones such as conversation, fiction, and YouTube vlog transcripts. The results indicate that the predictions of the mention detector (\texttt{Ours+MD}) provide improved mention detection scores, showing an increase of 1.4 points compared to the baseline model. Consequently, the model achieves an average F1 score of 66.4, outperforming the baseline model by 1.1 points. The improvement demonstrates the effectiveness of our proposed pipeline, which enhances the generalization on unseen data.

Because OntoGUM contains gold singleton annotation, we can directly use this mention information to assess the performance of the coreference clustering module within e2e systems. With gold singletons (\texttt{Ours+GS}), our pipeline achieves an average F1 score of 70.8, resulting in a larger improvement of 5.5 points over the baseline model (65.3). This improvement gap highlights the importance of gold singleton annotations. 

\subsection{Analysis}

\subsubsection{Qualitative Analysis}
We conduct a qualitative analysis comparing our mention predictions to the gold spans in OntoNotes. 

\paragraph{Recall}
The mention detector misses 4.36\% of coreference markables on ON-dev as in Table \ref{tab:md_perf}. We manually categorize these into five groups, as exemplified in Table \ref{tab:errors}. 
The first type of error relates to \texttt{missing nested entities}. For example, the nested and coordinated ``bridge-''modifier, ``Zhuhai - Hong Kong - Macao,'' and the second city ``Hong Kong'' are missing from predictions. The second type \texttt{attachment of prepositional phrases} also relates to complex NPs. Particularly, when a noun interacts with PP attachments, the mention detector occasionally fails to make accurate predictions. In addition, NPs are complex in \texttt{garden-path sentences}, challenging both the mention detector and human comprehension. Fourth, while \texttt{verbs} can have coreference relations in OntoNotes, they comprise a very small portion (less than 2\%) of the annotated data. Consequently, due to underrepresentation, the mention detector, which knows these only from the unioned gold coreference data, may miss some verbal markables. The remaining type is \texttt{gold annotation errors}. Coreference in OntoNotes relies heavily on syntax trees, particularly NP spans, and annotations in OntoNotes are not always correct. Some entities are incorrectly split into two parts due to annotation errors in the syntax tree. These splits are typically observed in post-nominal prepositional phrases and relative clauses. Additionally, redundant punctuations, such as extra commas and opening quotation marks, are sometimes incorrectly included.

The manual inspection reveals that recall errors are distributed sporadically, and a considerable subset of them cannot be avoided even with a better mention detector. Consequently, addressing these remains a substantial challenge and requires significant effort to bridge the gap.

\begin{table}[t!]
    \centering\small
    \begin{tabular}{p{0.45\textwidth}}
    \toprule
    \textbf{Recall} \\
    \hline
    \texttt{Missing nested entity}: Once the [Zhuhai - [Hong Kong] - Macao] bridge is built ... \\
    \texttt{Attachment of prepositional} \texttt{phrases}: He just told \underline{[a story] uh from the beginning to the end}. \\
    \texttt{Garden-path sentences}: Like \underline{[the bones]} \underline{xrays of his wisdom teeth} also tell us something about his age. \\
    \texttt{Missing verbal referents}: ... a unit of Marines [killed]\textsubscript{\#126} some 24 unarmed Iraqis ... [this atrocity]\textsubscript{\#126} ... \\
    \texttt{Gold annotation errors}: They can volunteer at \underline{[any] [of thousands of non-profit institutions]} ... \\
    \midrule
    \textbf{Precision} \\
    \hline
    \texttt{Redundant punctuations}: [one .] \\
    \texttt{Redundant non-restrictive relative clauses}: [5 p.m. EST -- when stocks there plunged.] \\
    \texttt{Generic NPs}: [no media] \\
    \bottomrule
    \end{tabular}
    \caption{Major categories of recall and precision errors in OntoNotes dev set. [Square brackets] denotes gold mention spans and \underline{underlining} indicates the most relevant predicted span (if necessary). Precision errors are enclosed by [square brackets].}
    \label{tab:errors}
\end{table}

\paragraph{Precision}
Due to the absence of singletons in OntoNotes annotations, we estimate that only $\sim$20\% of the missing precision
is relevant to mention detection errors (see Sec \ref{sec:training}).
Three types of precision errors were observed, as shown in Table \ref{tab:errors}. These errors include \texttt{redundant punctuations} as in ``one .'' or \texttt{redundant non-restrictive relative clauses} as in ``5 p.m. EST -- when stocks there plunged.''
The most tricky cases are \texttt{generic NPs} that do not refer to specific mentions. These include negative phrases such as 
``no media'' or quantifier phrases such as ``any of the Disney symbols.''
Though manually spotting such precision errors provides some insight, the lack of singleton annotations in OntoNotes hinders the possibility of exhaustively quantifying precision error types. 
This underscores the value of gold-singleton annotated corpora for mention detection evaluation.

\subsubsection{Effect of Mention Detection}

\begin{figure*}
\centering\small

\begin{minipage}{0.5\textwidth}
\centering
\begin{tikzpicture}[scale=0.8, transform shape]
\label{fig:recall}
\begin{axis}[
ylabel={Avg. F1},
xlabel={Recall},
xmin=95, xmax=100.5,
ymin=77, ymax=88,
xtick={95.5,96,97,98,99,100},
ytick={78,80,82,84,86,88},
legend pos=north west,
ylabel near ticks,
]

\addplot[
blue,
mark=triangle*,
mark size=3pt
]
coordinates {
(96,79.48)(97,80.03)(98,80.56)(99,81.24)(100,81.76)
};
\addplot[mark=*,blue,mark size=3pt]
coordinates {(95.64,79.27)};
\addplot[
style=dashed,
thick,
color=black,
]
coordinates {(95,79.4) (100.5,79.4)};

\legend{Avg. F1}

\end{axis}
\end{tikzpicture}
\end{minipage}%
\begin{minipage}{0.5\textwidth}
\centering
\begin{tikzpicture}[scale=0.8, transform shape]
\label{fig:precision}
\begin{axis}[
ylabel={Avg. F1},
xlabel={Precision},
xmin=32, xmax=105,
ymin=77, ymax=88,
xtick={35,40,50,60,70,80,90,100},
ytick={78,80,82,84,86,88},
legend pos=north west,
ylabel near ticks,
]

\addplot[
red,
mark=triangle*,
mark size=3pt
]
coordinates {
(40,79.94)(50,82.08)(60,83.32)(70,84.46)(80,85.02)(90,85.31)(100,85.64)
};
\addplot[mark=*,red,mark size=3pt]
coordinates {(38.74,79.27)};
\addplot[
style=dashed,
thick,
color=black,
]
coordinates {(32,79.4) (105,79.4)};
\draw [dashed, thick] (axis cs:80,77) -- (axis cs:80,88);

\legend{Avg. F1}

\end{axis}
\end{tikzpicture}
\end{minipage}

\caption{Analyzing the impact of recall and precision scores on the OntoNotes development set. The horizontal dashed line represents the baseline score and the rounded data point denotes the F1 scores achieved by the two-step training pipeline, aligned with their respective precision and recall scores. The vertical dashed line denotes an estimation of avg. F1 and precision score with gold singletons.}
\label{fig:pr}
\end{figure*}
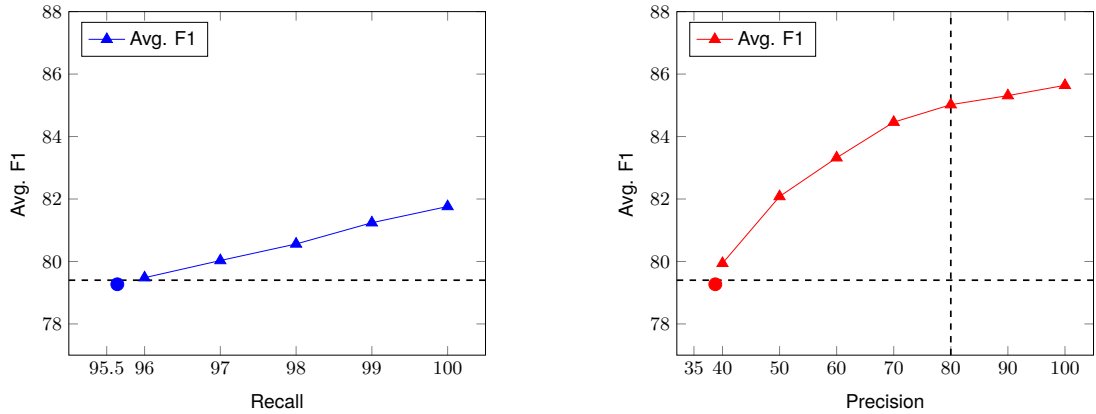

One of the advantages of the proposed pipeline is that the two separate steps provide more transparency than e2e models. Consequently, we investigate how mention detection affects coreference linking.

\paragraph{Recall}
The left side of Figure \ref{fig:pr} demonstrates the impact of recall on coreference resolution in the OntoNotes development set. We focus on the relationship between mention recall and coreference resolution performance.

Compared to precision, increasing the coverage of gold markables is a more intuitive way to benefit coreference clustering: markables that are undetected will inevitably lead to errors. Therefore, to investigate the significance of recall in coreference clustering, we explore different recall scenarios by adjusting the proportion of recall errors in the mention spans provided to the coreference module. 
We randomly omit certain portions of the recall errors to improve the recall score. These spans are evenly distributed among the documents in the OntoNotes development set. Initially, the mention detector's recall score stands at around 95.6\%, meaning it correctly identifies the majority of coreference markables. We then gradually increase the recall score from 96 to 100. $Recall=100$ indicates that all gold coreference markables are covered by predicted mentions.

The left panel in Figure \ref{fig:pr} also shows the average F1 score for various recall scores (number of mentions) within the OntoNotes development set. It exhibits a similar trend to the precision plot: with an increase in recall scores, the average F1 score also increases. When all recall errors are removed, the coreference score achieves an average F1 of 81.8, increasing the initial score by 2.4 points and outperforming the baseline model by 2.2 points, a substantial improvement but not as large as the maximal precision-based improvement.

\paragraph{Precision}
The results from Table \ref{tab:md_perf} demonstrate that the mention detector produces a substantial amount of precision errors, indicating that it identifies many potential mentions that are not a coreference markable. Such precision errors are unsurprising since singletons are excluded from coreference markables. Still, they can lead to the coreference resolution model making incorrect predictions, even if the model performs well in other respects.


To thoroughly explore how mention detection influences coreference clustering, like the recall plot, we modify the number of incorrect mentions from the mention detector's predictions and feed various mention spans to the coreference module, using the gold annotations to distinguish markables that will corefer with others (as shown in the right side of Figure \ref{fig:pr}). Specifically, we randomly eliminate incorrect markables from the OntoNotes development set to enhance the precision score. For instance, the initial precision score of the mention detector is 37.8 (as represented by the red circular data point), indicating that it retrieves around 48K mentions, of which 18K are accurate coreference markables. To increase the precision score to 50\%, we select 2.6K erroneous spans and distribute them equally among development documents. Starting from 37.8, the adjusted precision score ranges from 40 to 100. It is noted that $precision=100$ represents an ideal case,  indicating that all predicted mentions are gold coreference markables.

The right panel of Figure \ref{fig:pr} represents the average F1 score as a function of the precision score (number of mentions) within the OntoNotes development set. Unsurprisingly, higher mention precision scores correlate with enhanced average F1 scores. When we selectively remove 11\% of the precision errors ($\sim$5.6K) from the predicted mentions list by increasing the precision score from 39 (the circle point) to 50, we observe an average F1 score of 2.7 points increase, indicating a substantial improvement over the baseline model.

As discussed in Section \ref{sec:training}, we estimate around 40K mentions (including singletons) in the OntoNotes development set. Consequently, we can approach an upper bound on the precision score in our context. Given the number of predicted mentions (48K), we can, at most, randomly remove around 8K spans, resulting in an 80\% precision score (illustrated by the vertical dashed line in the figure). Given the current mention detector with nearly 96\% recall, the model can achieve the best coreference score, reaching an F1 of 85.0 with a precision of 80\%. This analysis demonstrates the critical role of mention precision in coreference resolution and its potential to impact model performance.

In sum, reducing both mention precision and recall errors substantially impacts coreference resolution performance. While the coreference clustering module can handle some incorrect spans by rejecting them during linking, certain precision errors continue to affect its performance. Furthermore, since existing systems already achieve high recall scores for mention detection, increasing recall further is challenging. By contrast, the precision score is still relatively low and can be improved more easily. Thus, focusing on precision improvement will likely offer more significant benefits for future coreference models.

\section{Conclusion}

This paper introduces a novel approach to address the coreference resolution challenge. It establishes a near-gold singleton dataset for OntoNotes, which is shown to be highly accurate. This dataset can benefit further research endeavors involving singletons in coreference systems. Additionally, we propose SPLICE, a pipeline-based neural system that independently trains mention detection and coreference models. Our system achieves comparable in-domain results with the e2e approach and demonstrates superior OOD performance. Leveraging the better interpretability of our system, we conduct a comprehensive analysis of mention predictions. We discover that resolving additional recall errors is more challenging than addressing precision errors, which offers valuable insight for future work in coreference resolution research.

\section{Limitations}
This work does not provide a manual validation of the performance of the singletons constructed by the NP classifier and the mention detector in OntoNotes, though we do provide an evaluation on the gold standard annotations in ARRAU.

The focus of this paper is on the NP to mention classification task for English datasets, and it might not cover certain linguistic phenomena found more often in other languages, such as zero anaphora in Chinese and Spanish. Furthermore, the performance of NP classification or mention detection aligning with data in other languages included in OntoNotes, such as Arabic \cite{pradhan-etal-2013-towards} or Chinese \cite{pradhan-etal-2013-towards}, and multiple languages in multilingual coreference benchmarks such as CorefUD \cite{nedoluzhko-etal-2022-corefud}, has not been evaluated in this work.

This paper evaluates the system's generalizability on the OntoGUM corpus. There are other challenging coreference datasets, such as GENTLE \cite{aoyama-etal-2023-gentle}, not included in our OOD evaluation.

\section{Bibliographical References}\label{sec:reference}

\bibliographystyle{lrec-coling2024-natbib}
\bibliography{anthology, custom}


\end{document}